\documentclass[runningheads]{llncs}
\usepackage[T1]{fontenc}
\usepackage{graphicx}
\usepackage{float}
\usepackage{tabularx}
\usepackage{array}
\usepackage{amsmath}
\usepackage{booktabs}
\usepackage{multirow}
\usepackage{amsfonts}
\begin{document}
\title{NeoTriFuse: Reliability-Aware Multimodal Fusion under Missingness Heterogeneity for Neonatal Mortality Risk Prediction}
\titlerunning{Reliability-Aware NeoTriFuse}
%
\author{Jiyuan Tian\inst{1} \and Qincheng Shen\inst{1} \and Ye Lin\inst{2} \and Yu Gao\inst{2} \and Haohui Lu\inst{2}}
\authorrunning{J. Tian et al.}
%
\institute{Discipline of Business Analytics, The University of Sydney, Sydney, Australia\\
\email{\{jtia0016,qshe0501\}@uni.sydney.edu.au}
\and
Molly Wardaguga Institute for First Nations Birth Rights, Charles Darwin University, Darwin, Australia\\
\email{\{haohui.lu,andrew.lin,yu.gao\}@cdu.edu.au}
}
\maketitle              
\begin{abstract}
Neonatal mortality risk prediction from bedside monitoring data remains challenging due to extreme class imbalance, heterogeneous clinical risk factors, multi-scale temporal dynamics, and substantial missingness. We propose NeoTriFuse, a reliability-aware multimodal fusion framework for missingness-heterogeneous neonatal monitoring data. Unlike conventional multimodal approaches that treat missingness primarily as a preprocessing issue, NeoTriFuse models missingness as an explicit reliability signal that dynamically modulates modality contributions during fusion. The framework integrates static perinatal variables, local–global temporal encoders, and patient-level statistical summaries through reliability-guided gating mechanisms, while jointly optimizing mortality prediction and an auxiliary length-of-stay objective. NeoTriFuse achieves competitive performance, with an F1 score of $0.6736 \pm 0.0216$ and an AUROC of $0.9454 \pm 0.0056$. Ablation studies indicate that the local–global temporal architecture and patient-level summary branch contribute most substantially to predictive performance, while reliability-aware gating provides additional improvements on threshold-dependent metrics under heterogeneous observation completeness. Sensitivity analyses further suggest stable performance across nearby hyperparameter settings. Overall, the findings support reliability-aware multimodal fusion as a practical approach for neonatal mortality prediction under realistic clinical missingness conditions.
\end{abstract}

\section{Introduction}
Risk assessment for neonatal deterioration is clinically valuable but still difficult in practice. Severe outcomes are rare, physiologic measurements are noisy and incomplete, and the same vital-sign pattern may have different meanings for infants with different gestational age, birth weight, or clinical history~\cite{richardson1998risk,mangold2021systematic}. In the Neonatal Intensive Care Unit (NICU), some risk appears as sudden short-term instability, while other risk is only visible as a slower change over several days. Prior work on heart-rate characteristics and HR/SpO$_2$ time-series features shows that repeated physiologic monitoring can carry prognostic information beyond baseline clinical variables~\cite{griffin2005hrc,niestroy2022fatal}. A mortality model therefore needs to read both the monitoring time series and the more stable perinatal context.

Recent neonatal studies suggest two lessons. Deep sequence models can learn patterns from vital-sign streams that are difficult to define manually~\cite{feng2021preterm}, while carefully designed clinical and statistical summaries can remain strong in small and imbalanced datasets~\cite{li2024survival,sullivan2025competition}. The Pediatric Academic Societies 2024 NICU Mortality Prediction Challenge illustrates this point: several teams used the same dataset, and simpler models remained competitive with more complex models when feature engineering and validation were handled carefully~\cite{sullivan2025competition,germanmesner2024pas}. This motivates a hybrid design that preserves explicit patient-level summaries while learning temporal representations from the monitoring stream.

The challenge is not only to maximize discrimination, but also to produce a clinically meaningful operating point. In rare mortality prediction, a model can obtain a high Area Under the Receiver Operating Characteristic Curve (AUROC) by ranking infants reasonably well while still producing weak precision-recall balance at a fixed threshold; precision-recall analysis is often more informative than ROC analysis when the positive class is rare~\cite{saito2015precision}. This distinction matters because clinical risk stratification depends on actionable risk groups and follow-up assessment rather than ranking alone, and clinical prediction guidance emphasizes threshold choice and validation rather than discrimination alone~\cite{steyerberg2019prediction}. At the same time, neonatal monitoring records are often heterogeneous: some infants have dense repeated measurements, whereas others have shorter or less complete trajectories. A useful model should therefore account for differences in observation reliability while still preserving simple patient-level summaries that are known to work well in tabular baselines.

This study asks whether neonatal mortality prediction can be improved by reliability-aware multimodal fusion under missingness heterogeneity. NeoTriFuse is designed to address this question by combining three forms of evidence: static perinatal risk, temporal vital dynamics, and patient statistical summaries. The goal is not to present architectural complexity as the main contribution, but to test whether missingness and coverage can be used as reliability signals that guide how these information sources are fused.

Our contributions are as follows:
\begin{itemize}
\item We formulate neonatal mortality prediction as reliability-aware multimodal fusion under missingness heterogeneity, where observation coverage helps determine how much each modality should contribute.
\item NeoTriFuse models missingness as an explicit reliability signal that dynamically modulates static, temporal, and summary representations rather than treating missingness solely as a preprocessing issue.
\item A local--global temporal architecture is introduced, combining convolutional layers for short-term physiological variations with transformer-based encoding for longer temporal dependencies.
\item Patient-level statistical summaries, including mean, variability, extrema, last observation, sequence length, and coverage, are preserved as a dedicated representation branch to retain clinically informative aggregate patterns.

\end{itemize}

\section{Related Work}
Traditional neonatal risk models use variables available around birth, including gestational age, birth weight, and early clinical condition, to estimate morbidity risk~\cite{richardson1998risk}. These variables remain important in recent machine-learning studies: Li et al.~\cite{li2024survival} used maternal and newborn admission features for survival prediction in extremely premature infants, and reviews emphasize cohort definition, validation, missing data handling, and interpretability~\cite{mangold2021systematic,kwok2022ai}. Static predictors are auditable and clinically meaningful, but they cannot fully describe deterioration that develops after admission. This motivates models that retain perinatal context while allowing repeated measurements to update the risk representation.

Continuous physiologic monitoring provides complementary information for neonatal risk prediction. Prior studies have linked heart-rate characteristics and HR/SpO$_2$ dynamics with neonatal infection, mortality, and respiratory complications~\cite{griffin2005hrc,niestroy2022fatal,qiu2024respiratory}. Deep learning approaches have also been applied to preterm mortality prediction and related ICU forecasting tasks~\cite{feng2021preterm}, while NICU sepsis studies highlight the importance of rigorous validation and comparison against simpler baselines when using repeated cardiorespiratory measurements~\cite{kausch2023sepsis,masino2019sepsis,reyna2019physionet}. Since neonatal deterioration may emerge through both short-term instability and longer-term temporal progression, NeoTriFuse incorporates both local temporal modelling and global attention-based representations rather than relying on a single temporal encoder.

The Pediatric Academic Societies (PAS) 2024 NICU Mortality Prediction Challenge is the closest shared-data reference for this study~\cite{germanmesner2024pas}. Sullivan et al.~\cite{sullivan2025competition} found that logistic regression, random forests, XGBoost, CatBoost, and neural network submissions can all be competitive when features and thresholds are chosen carefully. This agrees with clinical prediction guidance that emphasizes validation and task-appropriate evaluation over model complexity alone~\cite{james2021islr,steyerberg2019prediction}. Accordingly, our comparisons include gradient-boosted trees and random forests~\cite{prokhorenkova2018catboost,chen2016xgboost,ke2017lightgbm,feng2022trigger}, and our model preserves explicit patient-level summaries rather than relying only on learned sequence representations.

Finally, clinical monitoring data are shaped by missingness and coverage heterogeneity. Missing values may reflect workflow, monitoring duration, device availability, or clinical stability rather than random noise. Since simple imputation can discard this observation-reliability signal, NeoTriFuse treats missingness and coverage as inputs to reliability-aware fusion. This design is also compatible with the broader emphasis on inspectable clinical ML systems and post-hoc interpretation tools~\cite{lundberg2017shap}.

\section{Method}
\subsection{Problem Setting}
For each infant, let
\[
\mathbf{X}=[\mathbf{x}_1,\ldots,\mathbf{x}_T] \in \mathbb{R}^{T\times d_t}
\]
denote the temporal monitoring sequence, where each time step contains HR/SpO$_2$ summary variables, timing features, and explicit missingness indicators. In this benchmark formulation, $\mathbf{X}$ contains all available monitoring records for each infant in the model data table. The task is therefore retrospective full-record mortality risk classification at the end of the available monitoring record, rather than fixed-horizon early warning at admission or within the first 24--48 hours. Let $\mathbf{s}\in\mathbb{R}^{d_s}$ denote static demographic and perinatal variables, and let $y\in\{0,1\}$ be the in-hospital mortality label. In addition to mortality prediction, we use the log-transformed length of stay
\[
\ell=\log(1+\mathrm{LOS})
\]
as an auxiliary regression target. The model estimates both the mortality probability $\hat{y}$ and the auxiliary LOS value $\hat{\ell}$ from $(\mathbf{X},\mathbf{s})$.

\subsection{Input Representation}
The static vector $\mathbf{s}$ includes demographic and perinatal covariates such as gestational age, birth weight, sex, Apgar scores, delivery mode, and maternal-age-related information. Missing temporal values are not simply discarded. Instead, the input includes mask features so that the model can distinguish an observed normal value from an absent value.

To keep stable patient-level information available to the neural model, we also compute a summary vector
\[
\mathbf{z}=\psi(\mathbf{X})\in\mathbb{R}^{d_z},
\]
where $\psi(\cdot)$ contains feature-wise mean, standard deviation, minimum, maximum, last observed value, sequence length, and coverage ratio. These features are close to the summaries used by strong tabular baselines, but here they are learned jointly with the temporal branches.

This representation deliberately separates raw temporal evidence from aggregate patient descriptors. If the summary features were only appended to the temporal sequence, the model would need to learn their patient-level meaning indirectly. If they were used only by a separate tree-based baseline, the neural model would lose access to robust descriptors that are especially useful in small clinical datasets. By making $\mathbf{z}$ a dedicated branch, NeoTriFuse can combine stable summary information with learned temporal patterns while still allowing ablation to test whether the branch adds value.

\subsection{NeoTriFuse}
\begin{figure}[t]
\centering
\includegraphics[width=0.99\textwidth]{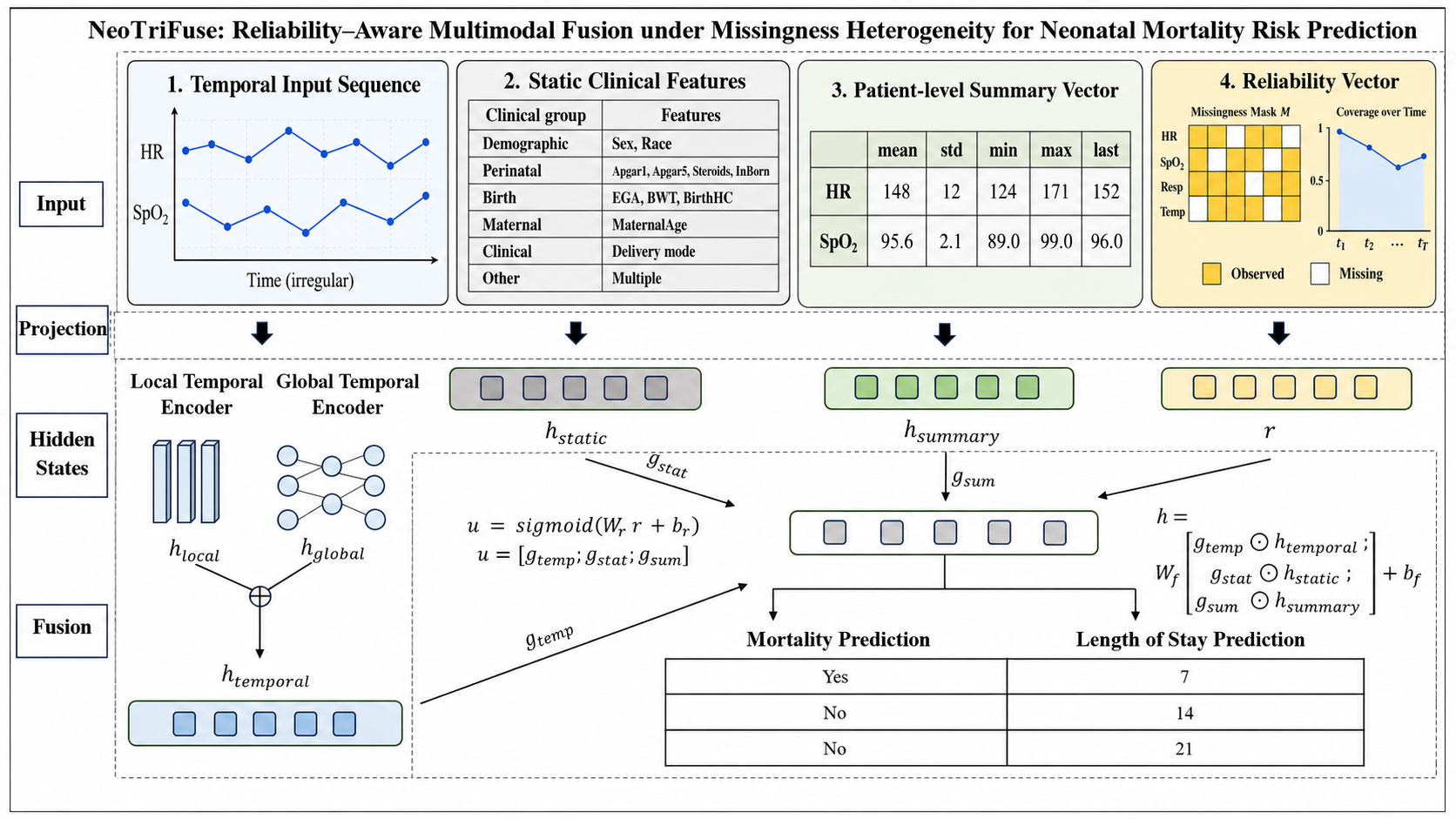}
\caption{Overview of NeoTriFuse for reliability-aware multimodal fusion under heterogeneous missingness. The model integrates temporal HR and SpO$_2$ sequences, static clinical variables, patient-level summary statistics, and a reliability vector derived from missingness and observation coverage. Local and global temporal encoders generate $\mathbf{h}_{temporal}$, while reliability-guided gates modulate temporal, static, and summary representations before fusion. The fused representation is used for mortality prediction with auxiliary length-of-stay prediction.}

\label{fig:architecture}
\end{figure}

Figure~\ref{fig:architecture} provides an overview of NeoTriFuse. The model is organized around reliability-aware multimodal fusion rather than simple feature concatenation: three input branches provide complementary evidence, and missingness-derived reliability gates modulate their contributions before prediction.

\subsubsection{Local--Global Temporal Encoding}
The local branch applies one-dimensional temporal convolutions to capture short changes in the monitoring sequence:
\[
\mathbf{h}_{loc}=\mathrm{Pool}(\mathrm{CNN}(\mathbf{X})).
\]
The global branch uses a transformer encoder to model longer-range dependencies~\cite{vaswani2017attention}:
\[
\mathbf{h}_{glob}=\mathrm{Pool}(\mathrm{Transformer}(\mathbf{X})).
\]
The two temporal representations are concatenated and projected to obtain
\[
\mathbf{h}_{temp}=W_t[\mathbf{h}_{loc};\mathbf{h}_{glob}]+\mathbf{b}_t .
\]
This local--global design is used because neonatal deterioration may appear either as abrupt instability or as a slower trend across the monitoring window.

\subsubsection{Static and Summary Branches}
The static clinical variables and the summary vector are mapped into the same hidden dimension:
\[
\mathbf{h}_{stat}=f_s(\mathbf{s}), \qquad
\mathbf{h}_{sum}=f_z(\mathbf{z}),
\]
where $f_s$ and $f_z$ are multilayer perceptrons. Keeping $\mathbf{h}_{sum}$ as a separate branch avoids forcing the temporal encoder to rediscover simple but useful descriptors such as coverage, extrema, and last observation.

\subsubsection{Reliability-Aware Fusion}
We compute a reliability vector $\mathbf{r}$ from sequence coverage and missingness-derived completeness. Let $v_t\in\{0,1\}$ indicate whether time step $t$ contains a valid temporal observation, let $m^s_j\in\{0,1\}$ indicate whether static feature $j$ is missing, and let $m^z_k\in\{0,1\}$ indicate whether summary feature $k$ is missing. The reliability vector is defined as
\[
c_{temp}=\frac{1}{T}\sum_{t=1}^{T}v_t,\qquad
c_{stat}=1-\frac{1}{d_s}\sum_{j=1}^{d_s}m^s_j,\qquad
c_{sum}=1-\frac{1}{d_z}\sum_{k=1}^{d_z}m^z_k,
\]
\[
\mathbf{r}=[c_{temp};c_{stat};c_{sum}].
\]
Here $c_{temp}$ measures temporal coverage, while $c_{stat}$ and $c_{sum}$ measure the observed-feature completeness of the static and summary branches. This vector is used to generate branch-wise gates:
\[
\mathbf{u}=\sigma(W_r\mathbf{r}+\mathbf{b}_r), \qquad
\mathbf{u}=[\mathbf{g}_{temp};\mathbf{g}_{stat};\mathbf{g}_{sum}].
\]
The fused representation is then
\[
\mathbf{h}=[\mathbf{g}_{temp}\odot\mathbf{h}_{temp};
\mathbf{g}_{stat}\odot\mathbf{h}_{stat};
\mathbf{g}_{sum}\odot\mathbf{h}_{sum}],
\]
followed by a shared projection layer. This mechanism allows the model to reduce reliance on a branch when the corresponding information is sparse or unreliable.

The reliability vector should be interpreted as a data quality signal rather than a direct clinical risk factor. For example, a short or sparsely observed temporal record may make dynamic features less dependable, even if the static variables remain informative. In contrast, a densely observed record may allow the temporal branch to carry stronger evidence about recent instability. The gates provide a simple way to let this reliability information modulate the multimodal representation without requiring a separate model for missingness pattern.

\subsubsection{Prediction Heads and Objective}
The mortality head predicts $\hat{y}=\sigma(f_y(\mathbf{h}))$, and the auxiliary head predicts $\hat{\ell}=f_\ell(\mathbf{h})$. Mortality prediction uses focal binary cross-entropy with positive-class weighting~\cite{lin2017focal}. For a sample with label $y\in\{0,1\}$ and predicted probability $p$, define
\[
p_t = yp + (1-y)(1-p), \qquad
\alpha_t = y\alpha + (1-y)(1-\alpha).
\]
The mortality loss is
\[
\mathcal{L}_{mortality}
= -\alpha_t(1-p_t)^{\gamma}
\left[w_+y\log(p) + (1-y)\log(1-p)\right],
\]
where $\alpha=0.75$ and $\gamma=2.0$ are fixed across all neural experiments. The positive-class weight is computed within each training fold as
\[
w_+=\frac{N_-}{\max(N_+,1)},
\]
where $N_+$ and $N_-$ denote the numbers of mortality-positive and mortality-negative infants in the training split. This fold-specific weighting avoids using validation-label information while compensating for severe mortality imbalance. We do not tune $\alpha$ or $\gamma$; they are kept fixed to avoid adding another validation-dependent hyperparameter search.

LOS prediction uses SmoothL1 loss. The total loss is
\[
\mathcal{L} = \mathcal{L}_{mortality} + \lambda_{los}\mathcal{L}_{los}.
\]
The auxiliary LOS target provides dense supervision for all patients, including the majority who survive, and regularizes the shared representation under severe mortality imbalance.

The auxiliary objective is used only as a training signal and is not the primary task of the study. Its role is to encourage the shared representation to encode clinically relevant severity information even when the mortality label is sparse. This is useful because length of stay is observed for many more patients than the positive mortality outcome and can provide a smoother gradient during optimization. The LOS weight therefore controls a tradeoff: a small value can regularize the representation, while an overly large value may shift the model away from the classification objective. This is the reason the sensitivity study explicitly varies $\lambda_{los}$ rather than treating it as a fixed implementation detail.

\section{Experiments}
\subsection{Research Questions}

The experimental evaluation is guided by the following research questions:
\begin{itemize}
\item RQ1: How does NeoTriFuse perform relative to standard tabular and sequence-based baselines for neonatal mortality prediction?
\item RQ2: What is the contribution of the main architectural components, including local temporal encoding, global temporal encoding, patient-level summaries, reliability-aware fusion, and the auxiliary LOS prediction task?
\item RQ3: Does the model maintain stable performance under moderate changes to the hidden dimension size and LOS loss weighting?
\end{itemize}

\subsection{Dataset and Split}
We use the PAS 2024 NICU Mortality Prediction Challenge dataset released through the University of Virginia Dataverse~\cite{germanmesner2024pas}. The cohort contains patient-level demographic information, outcome labels, and sequential HR/SpO$_2$ summary windows. In the current formulation, mortality is evaluated at the patient level using the official fold assignments and all available monitoring windows for each infant. The study focuses on the subset of variables that can be aligned reliably across all patients: serial HR/SpO$_2$ summary statistics from the model data table, demographic and perinatal variables from the demographic table, and mortality plus length-of-stay labels from the outcome table. All preprocessing is performed within the official fold protocol. Temporal windows are aligned at the patient level, and derived summary statistics are computed from the available sequence for each infant. Static variables and temporal summaries are then joined using patient identifiers. Because the summaries are computed over the full available record, the reported task should not be interpreted as prospective early-warning prediction. This setup keeps the comparison focused on modeling choices rather than on alternative data splits or extra external information.

\subsection{Experimental Setup}
The official cross-validation folds are used and four classification metrics are reported: F1, precision, recall, and AUROC. The data split is fixed by the official fold assignments for all runs; random seeds affect only model initialization, minibatch ordering, dropout, and other stochastic training operations. Comparison and ablation experiments are repeated across 10 random seeds, and we report mean and standard deviation.

\subsection{Implementation Details}
All neural models are trained with AdamW. NeoTriFuse uses focal BCE for mortality and SmoothL1 loss for LOS, along with validation threshold selection to improve the precision and recall tradeoff. We use the same fold protocol across all methods so that performance differences are attributable to modeling choices rather than data partition changes. 

The threshold used for reporting F1, precision, and recall is selected on validation predictions rather than fixed at 0.5, because mortality is a rare outcome and the default probability threshold can be poorly matched to the desired operating point. AUROC is reported separately because it evaluates ranking performance independent of a single threshold. Reporting both types of metrics helps distinguish models that rank patients well from models that also produce a useful classification balance.

\subsection{Baselines}
We compare against traditional machine-learning baselines, including Logistic Regression and Random Forest~\cite{james2021islr}, XGBoost~\cite{chen2016xgboost}, LightGBM~\cite{ke2017lightgbm}, and CatBoost~\cite{prokhorenkova2018catboost}. We also include neural sequence baselines, including GRU~\cite{cho2014gru}, CNN~\cite{wang2017time}, and Transformer~\cite{vaswani2017attention}. To address missingness-aware and multimodal temporal comparison, we further include GRU-D~\cite{che2018grud}, TNformer-MP~\cite{xu2024tnformermp}, and TIM-MSFL~\cite{liu2025timmsfl}; TIM-MSFL is treated as the multiscale temporal fusion baseline. 

\subsection{Ablation and Sensitivity Settings}
The ablation study removes the local CNN, global transformer, tabular summary branch, reliability-aware fusion mechanism, reliability vector $\mathbf{r}$, and LOS auxiliary objective in turn. The No Reliability Vector setting keeps the gating and fusion structure but replaces the reliability embedding derived from coverage and missingness with zeros, so that $\mathbf{r}$ cannot modulate branch contributions. Each ablation keeps the remaining training protocol unchanged so that performance differences reflect the removed component rather than changes in optimization or evaluation. Hyperparameter sensitivity evaluates hidden dimensions of 64, 96, and 128 with LOS-loss weights of 0.1, 0.2, and 0.3 across 10 random seeds. The sweep is deliberately local because its purpose is not exhaustive tuning; instead, it checks whether performance remains stable across nearby capacity and auxiliary-loss settings.

\section{Results}
\subsection{Comparison Analysis}
\begin{figure}[ht]
\centering
\includegraphics[width=0.9\textwidth]{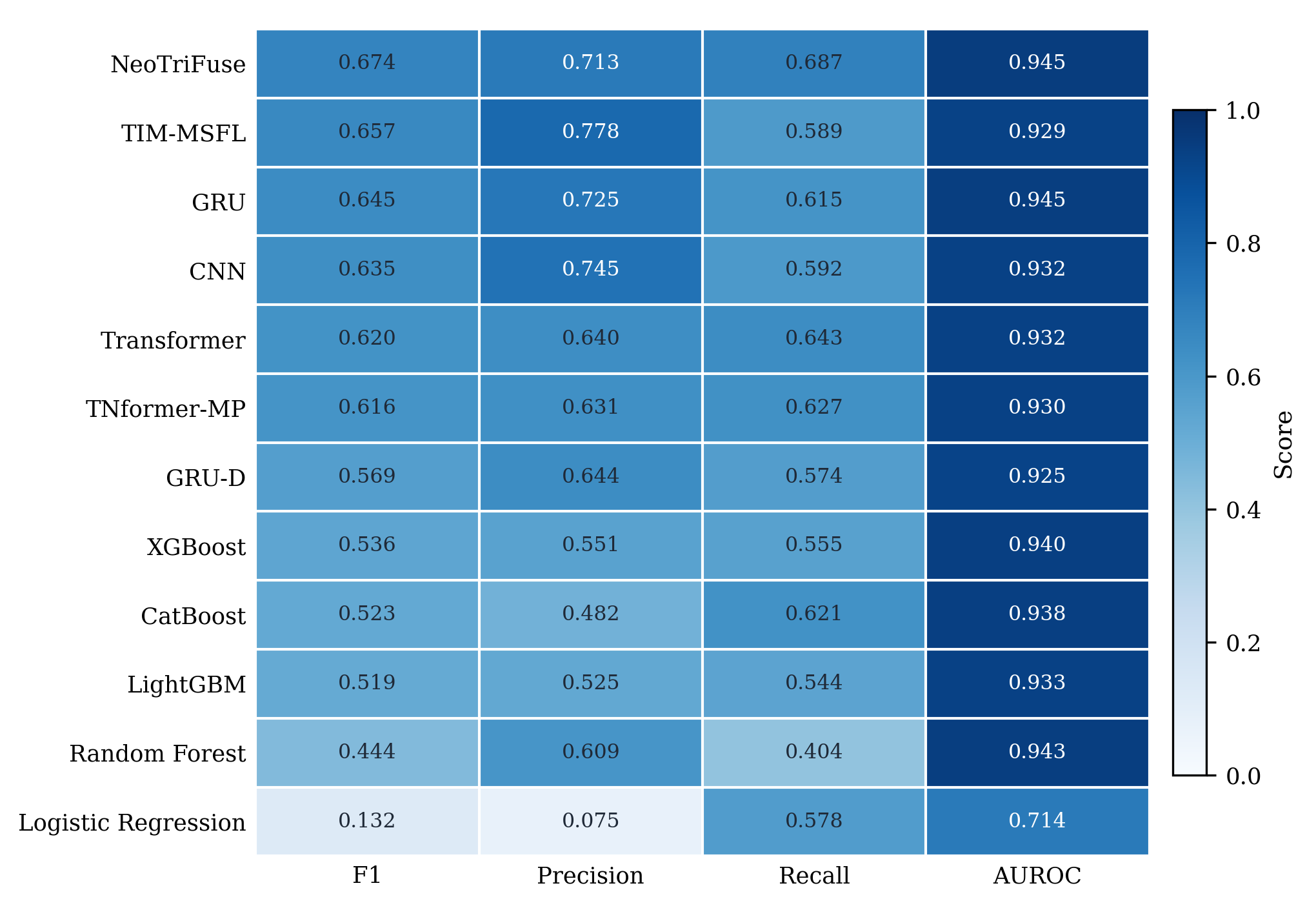}
\caption{Heatmap of classification performance across baselines and the proposed model.}
\label{fig:comparison}
\end{figure}

Table~\ref{tab:comparison} shows NeoTriFuse achieves the highest F1, recall, and AUROC among the evaluated models, including traditional machine-learning, neural sequence, and missingness-aware baselines. Figure~\ref{fig:comparison} further visualizes the metric-level pattern as a model and metric heatmap, making the contrast between threshold-dependent measures and ranking performance explicit. The strongest competing baseline is TIM-MSFL, which attains the highest precision but lower recall and F1. Compared with the strongest traditional baseline, XGBoost, NeoTriFuse improves F1 by more than 0.13 absolute points while also retaining higher AUROC.

The metric profile is also clinically meaningful. Several baselines obtain competitive AUROC or precision but weaker recall-balanced F1, indicating weaker classification balance under the selected operating threshold. By contrast, NeoTriFuse improves not only ranking quality but also the balance between precision and recall. This matters for clinical risk stratification, where a model that only sorts patients well but performs poorly at the chosen threshold can produce less useful risk groups.
\begin{table}[ht]
\caption{Comparison with traditional, neural, and missingness-aware baselines. Values are mean $\pm$ std over 10 seeds.}
\label{tab:comparison}
\centering
\small
\begin{tabular}{lcccc}
\toprule
Model & F1 & Precision & Recall & AUROC \\
\midrule
TIM-MSFL & 0.6570$\pm$0.0126 & \textbf{0.7780$\pm$0.0370} & 0.5888$\pm$0.0170 & 0.9292$\pm$0.0014 \\
GRU & 0.6446$\pm$0.0117 & 0.7246$\pm$0.0421 & 0.6147$\pm$0.0245 & 0.9450$\pm$0.0019 \\
CNN & 0.6348$\pm$0.0120 & 0.7445$\pm$0.0310 & 0.5916$\pm$0.0183 & 0.9317$\pm$0.0042 \\
Transformer & 0.6201$\pm$0.0164 & 0.6397$\pm$0.0314 & 0.6430$\pm$0.0211 & 0.9316$\pm$0.0037 \\
TNformer-MP & 0.6155$\pm$0.0202 & 0.6314$\pm$0.0357 & 0.6273$\pm$0.0205 & 0.9303$\pm$0.0039 \\
GRU-D & 0.5690$\pm$0.0223 & 0.6442$\pm$0.0438 & 0.5739$\pm$0.0234 & 0.9250$\pm$0.0030 \\
XGBoost & 0.5356$\pm$0.0127 & 0.5514$\pm$0.0104 & 0.5554$\pm$0.0182 & 0.9402$\pm$0.0021 \\
CatBoost & 0.5231$\pm$0.0122 & 0.4819$\pm$0.0147 & 0.6213$\pm$0.0082 & 0.9379$\pm$0.0031 \\
LightGBM & 0.5193$\pm$0.0073 & 0.5253$\pm$0.0099 & 0.5445$\pm$0.0054 & 0.9328$\pm$0.0015 \\
Random Forest & 0.4443$\pm$0.0175 & 0.6088$\pm$0.0204 & 0.4038$\pm$0.0171 & 0.9433$\pm$0.0022 \\
Logistic Regression & 0.1319$\pm$0.0000 & 0.0754$\pm$0.0000 & 0.5780$\pm$0.0000 & 0.7143$\pm$0.0000 \\
NeoTriFuse  & \textbf{0.6736$\pm$0.0216} & 0.7125$\pm$0.0266 & \textbf{0.6870$\pm$0.0201} & \textbf{0.9454$\pm$0.0056} \\
\bottomrule
\end{tabular}
\end{table}

\subsection{Ablation and Sensitivity Analysis}
Table~\ref{tab:ablation} shows that the complete model provides the best F1 balance among the tested configurations. Removing the global transformer branch causes the largest F1 degradation, followed by removing the tabular summary branch. This supports our central claim that strong patient-level summary statistics and long-range temporal context are both necessary. Replacing cross-gating with simple concatenation also reduces F1, although the gap is smaller.

Three ablation findings are particularly informative. First, removing the global transformer harms performance more than removing the local CNN, suggesting that longer-range trajectory context is especially important for mortality prediction in this cohort. Second, removing the tabular summary branch produces a clear F1 drop despite leaving the temporal branch intact. This indicates that explicit summary statistics are not redundant; instead, they provide complementary patient-level information that the sequence encoder alone does not recover reliably. Third, removing the reliability vector keeps AUROC high but lowers F1 and recall relative to the full model, suggesting that $\mathbf{r}$ mainly helps the threshold-dependent decision balance rather than ranking alone. The relatively small but consistent drop from replacing cross-gating with simple concatenation further suggests that adaptive interaction is useful even when the summary and temporal branches are already strong.

\begin{figure}[ht]
\centering
\includegraphics[width=0.99\textwidth]{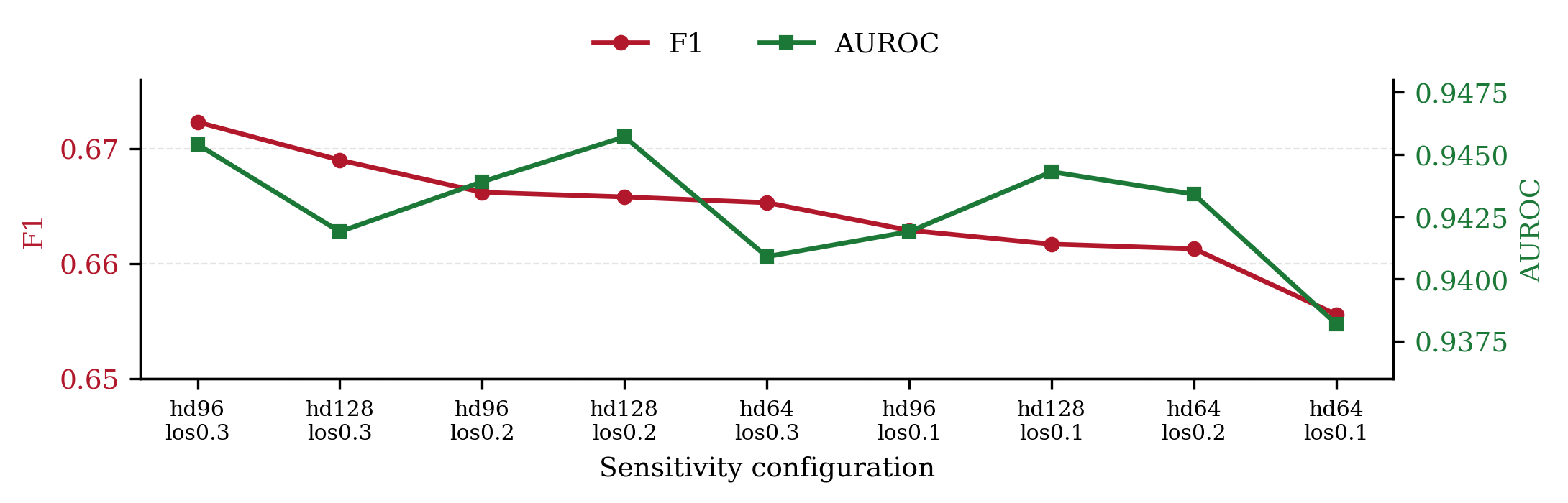}
\caption{Sensitivity of F1 and AUROC across hidden dimension and LOS loss weight settings.}
\label{fig:sensitivity_line}
\end{figure}

The 10-seed sensitivity sweep in Table~\ref{tab:sensitivity} and Figure~\ref{fig:sensitivity_line} examines whether the model is overly dependent on a narrow hyperparameter choice. The highest mean F1 is observed at hidden dimension 96 and LOS loss weight 0.3, but several nearby settings remain close in F1 and AUROC. Across the nine tested configurations, mean F1 ranges from 0.6556 to 0.6736, and mean AUROC ranges from 0.9382 to 0.9457. This indicates that the model remains reasonably stable across nearby capacity and auxiliary-loss settings, although small differences among configurations should be interpreted descriptively rather than as evidence of a uniquely optimal setting.

\begin{table}[h]
\caption{Ablation study of the proposed model.}
\label{tab:ablation}
\centering
\small
\resizebox{\textwidth}{!}{%
\begin{tabular}{lcccc}
\toprule
Variant & F1 & Precision & Recall & AUROC \\
\midrule
No Local CNN & 0.6632$\pm$0.0095 & 0.6961$\pm$0.0319 & 0.6824$\pm$0.0210 & 0.9450$\pm$0.0038 \\
No Global Transformer & 0.6412$\pm$0.0171 & 0.6633$\pm$0.0381 & 0.6750$\pm$0.0267 & 0.9292$\pm$0.0078 \\
No Tabular Summary & 0.6470$\pm$0.0115 & 0.6949$\pm$0.0332 & 0.6633$\pm$0.0232 & 0.9420$\pm$0.0080 \\
Concat Fusion & 0.6679$\pm$0.0090 & \textbf{0.7427$\pm$0.0309} & 0.6539$\pm$0.0213 & 0.9436$\pm$0.0048 \\
No Reliability Vector & 0.6685$\pm$0.0127 & 0.7157$\pm$0.0513 & 0.6663$\pm$0.0335 & \textbf{0.9473$\pm$0.0052} \\
No LOS Auxiliary & 0.6641$\pm$0.0156 & 0.7076$\pm$0.0588 & 0.6735$\pm$0.0309 & 0.9372$\pm$0.0045 \\
Full NeoTriFuse & \textbf{0.6736$\pm$0.0216} & 0.7125$\pm$0.0266 & \textbf{0.6870$\pm$0.0201} & 0.9454$\pm$0.0056 \\
\bottomrule
\end{tabular}
}
\end{table}

\begin{table}[h]
\caption{Ten-seed sensitivity analysis of hidden dimension and LOS loss weight.}
\label{tab:sensitivity}
\centering
\small
\begin{tabular}{lcccc}
\toprule
Config & F1 & Precision & Recall & AUROC \\
\midrule

hd128\_los0.3 & 0.6690$\pm$0.0162 & 0.7144$\pm$0.0370 & 0.6778$\pm$0.0233 & 0.9419$\pm$0.0037 \\
hd96\_los0.2 & 0.6662$\pm$0.0164 & 0.6999$\pm$0.0403 & 0.6824$\pm$0.0252 & 0.9439$\pm$0.0060 \\
hd128\_los0.2 & 0.6658$\pm$0.0144 & 0.6945$\pm$0.0414 & 0.6785$\pm$0.0329 & \textbf{0.9457$\pm$0.0044} \\
hd64\_los0.3 & 0.6653$\pm$0.0142 & 0.7218$\pm$0.0573 & 0.6653$\pm$0.0424 & 0.9409$\pm$0.0058 \\
hd96\_los0.1 & 0.6629$\pm$0.0204 & 0.7060$\pm$0.0407 & 0.6711$\pm$0.0307 & 0.9419$\pm$0.0079 \\
hd128\_los0.1 & 0.6617$\pm$0.0128 & 0.7112$\pm$0.0336 & 0.6667$\pm$0.0237 & 0.9443$\pm$0.0049 \\
hd64\_los0.2 & 0.6613$\pm$0.0091 & \textbf{0.7240$\pm$0.0400} & 0.6549$\pm$0.0335 & 0.9434$\pm$0.0026 \\
hd64\_los0.1 & 0.6556$\pm$0.0159 & 0.7092$\pm$0.0390 & 0.6561$\pm$0.0302 & 0.9382$\pm$0.0084 \\
hd96\_los0.3 & \textbf{0.6736$\pm$0.0216} & 0.7125$\pm$0.0266 & \textbf{0.6870$\pm$0.0201} &  0.9454$\pm$0.0056 \\
\bottomrule
\end{tabular}
\end{table}

\subsection{Comparison with Related Studies}
\label{sec:related_studies}

\begin{table}[h]
\centering
\small

\caption{Comparison with related neonatal mortality and NICU survival prediction studies.}
\label{tab:literature_comparison}
\renewcommand{\arraystretch}{1.05}
\setlength{\tabcolsep}{2pt}
\small
\begin{tabular}{@{}
>{\raggedright\arraybackslash}p{0.15\textwidth}
>{\raggedright\arraybackslash}p{0.20\textwidth}
>{\raggedright\arraybackslash}p{0.20\textwidth}
>{\raggedright\arraybackslash}p{0.20\textwidth}
>{\raggedright\arraybackslash}p{0.21\textwidth}@{}}
\toprule
\textbf{Model} & \textbf{Cohort and input data} & \textbf{Method} & \textbf{Results} & \textbf{Comment} \\
\midrule

Feng et al.~\cite{feng2021preterm} &
285 low birth weight preterm infants; clinical variables and vital signs &
Deep learning  &
\begin{tabular}[t]{@{}l@{}}
ACC: 0.888 \\
AUROC: 0.897 \\
Recall: 0.780
\end{tabular} &
Strong temporal model; small cohort \\

\addlinespace[0.3em]

Niestroy et al.~\cite{niestroy2022fatal} &
5,957 NICU infants; daily 10-min HR and SpO$_2$ segments &
Time-series features with logistic regression &
AUROC: 0.828 &
Effective feature engineering \\

\addlinespace[0.3em]

Li et al.~\cite{li2024survival} &
459 MIMIC-III infants, 23--29 weeks; admission features &
Machine learning models &
\begin{tabular}[t]{@{}l@{}}
AUROC: 0.91 \\
F1: 0.67
\end{tabular} &
Admission-based prediction \\

\addlinespace[0.3em]

Sullivan et al.~\cite{sullivan2025competition} &
PAS challenge data; demographics and HR/SpO$_2$ samples &
Machine learning models &
AUROC: 0.818 &
Strong PAS baseline \\

\addlinespace[0.3em]

\textbf{NeoTriFuse} &
PAS challenge data; multimodal inputs including static variables, HR/SpO$_2$, and masks &
Reliability-aware multimodal fusion &
\begin{tabular}[t]{@{}l@{}}
AUROC: 0.9454 \\
F1: 0.6736
\end{tabular} &
Missingness as reliability signal \\

\bottomrule
\end{tabular}
\end{table}

Table~\ref{tab:literature_comparison} compares NeoTriFuse with representative neonatal mortality and NICU survival prediction studies. The listed studies differ in cohort size, input modality, prediction window, and validation design, so the table should be interpreted as contextual evidence rather than a strict comparison. Even with this caveat, the comparison highlights three useful patterns. First, prior deep vital-sign models such as Feng et al. achieved strong accuracy and recall in a smaller single center cohort, showing the value of temporal physiology but leaving open the question of broader challenge-style validation. Second, Niestroy et al. and Sullivan et al. used PAS-related HR/SpO$_2$ and demographic information and reported AUROC value of 0.828, indicating that explicit time-series summaries and simple baselines are already competitive in this setting. Third, Li et al. reported a strong random-forest result on MIMIC-III admission variables, but that task focuses on initial survival prediction rather than longitudinal vital-sign modeling.

\section{Conclusion}
NeoTriFuse frames neonatal mortality prediction as reliability-aware multimodal fusion under missingness heterogeneity. By using missingness and coverage to gate temporal, static, and summary representations, the model offers a compact way to combine physiologic trajectories with stable patient-level descriptors while preserving observation reliability as an explicit signal.

This study remains limited by its retrospective full-record benchmark setting, the absence of matched external reimplementation against prior neonatal studies, and the limited availability of independent cohorts with the same multimodal inputs and fold definitions. Future work should evaluate fixed prospective prediction horizons, external validation, and clinical interpretation of the learned temporal and summary representations before considering real-time clinical use.

\begin{credits}

\subsubsection{\discintname}
The authors have no competing interests to declare that are relevant to the content of this article.
\end{credits}
%
%
%
\bibliographystyle{splncs04}
\bibliography{references}

@article{niestroy2022fatal,
  title={Discovery of signatures of fatal neonatal illness in vital signs using highly comparative time-series analysis},
  author={Niestroy, Justin C. and Moorman, J. Randall and Levinson, Maxwell A. and Al Manir, Sadnan and Clark, Timothy W. and Fairchild, Karen D. and Lake, Douglas E.},
  journal={npj Digital Medicine},
  volume={5},
  pages={6},
  year={2022},
  doi={10.1038/s41746-021-00551-z}
}

@article{feng2021preterm,
  title={Predicting mortality risk for preterm infants using deep learning models with time-series vital sign data},
  author={Feng, Jiarui and Lee, Jennifer and Vesoulis, Zachary A. and Li, Fuhai},
  journal={npj Digital Medicine},
  volume={4},
  pages={108},
  year={2021},
  doi={10.1038/s41746-021-00479-4}
}

@article{li2024survival,
  title={Improving prediction of survival for extremely premature infants born at 23 to 29 weeks gestational age in the neonatal intensive care unit: development and evaluation of machine learning models},
  author={Li, Angie and Mullin, Sarah and Elkin, Peter L.},
  journal={JMIR Medical Informatics},
  volume={12},
  pages={e42271},
  year={2024},
  doi={10.2196/42271}
}

@article{sullivan2025competition,
  title={Comparing machine learning techniques for neonatal mortality prediction: insights from a modeling competition},
  author={Sullivan, Brynne A. and Moreira, Alvaro G. and McAdams, Ryan M. and Knake, Lindsey A. and Husain, Ameena and Qiu, Jiaxing and Mudireddy, Avinash and Majeedi, Abrar and Shalish, Wissam and Lake, Douglas E. and Vesoulis, Zachary A.},
  journal={Pediatric Research},
  volume={98},
  pages={405--411},
  year={2025},
  doi={10.1038/s41390-024-03773-5}
}

@misc{germanmesner2024pas,
  title={{Pediatric Academic Societies} 2024 {NICU} Mortality Prediction Challenge},
  author={German Mesner, Ian},
  year={2024},
  howpublished={University of Virginia Dataverse},
  doi={10.18130/V3/5UYB4U}
}

@article{richardson1998risk,
  title={Neonatal risk scoring systems: Can they predict mortality and morbidity?},
  author={Richardson, Douglas K. and Tarnow-Mordi, William O. and Escobar, Gabriel J.},
  journal={Clinics in Perinatology},
  volume={25},
  number={3},
  pages={591--608},
  year={1998},
  doi={10.1016/S0095-5108(18)30099-X}
}

@article{mangold2021systematic,
  title={Machine learning models for predicting neonatal mortality: a systematic review},
  author={Mangold, Carina and Zoretic, Samuel and Thallapureddy, Karthik and Moreira, Alvaro and Chorath, Kishore and Moreira, Arvind},
  journal={Neonatology},
  volume={118},
  pages={394--405},
  year={2021},
  doi={10.1159/000516891}
}

@article{kwok2022ai,
  title={Application and potential of artificial intelligence in neonatal medicine},
  author={Kwok, T'ng Chang and Henry, Caroline and Saffaran, Sina and Meeus, Marisse and Bates, Declan and Van Laere, David and Boylan, Geraldine and Boardman, James P. and Sharkey, Don},
  journal={Seminars in Fetal and Neonatal Medicine},
  volume={27},
  number={5},
  pages={101346},
  year={2022},
  doi={10.1016/j.siny.2022.101346}
}

@article{griffin2005hrc,
  title={Heart rate characteristics: novel physiomarkers to predict neonatal infection and death},
  author={Griffin, M. Pamela and Lake, Douglas E. and Bissonette, Eric A. and Harrell, Frank E. and O'Shea, T. Michael and Moorman, J. Randall},
  journal={Pediatrics},
  volume={116},
  number={5},
  pages={1070--1074},
  year={2005},
  doi={10.1542/peds.2004-2461}
}

@article{kausch2023sepsis,
  title={Cardiorespiratory signature of neonatal sepsis: development and validation of prediction models in 3 NICUs},
  author={Kausch, Sherry L. and Brandberg, Julia G. and Qiu, Jiaxing and Panda, Arka and Binai, Abtin and Isler, Jennifer and Sahni, Rakesh and Vesoulis, Zachary A. and Moorman, J. Randall and Fairchild, Karen D. and Lake, Douglas E. and Sullivan, Brynne A.},
  journal={Pediatric Research},
  volume={93},
  pages={1913--1921},
  year={2023},
  doi={10.1038/s41390-022-02444-7}
}

@article{masino2019sepsis,
  title={Machine learning models for early sepsis recognition in the neonatal intensive care unit using readily available electronic health record data},
  author={Masino, Aaron J. and Harris, Mary Catherine and Forsyth, Daniel and Ostapenko, Svetlana and Srinivasan, Lakshmi and Bonafide, Christopher P. and Balamuth, Fran and Schmatz, Melissa and Grundmeier, Robert W.},
  journal={PLOS ONE},
  volume={14},
  number={2},
  pages={e0212665},
  year={2019},
  doi={10.1371/journal.pone.0212665}
}

@article{qiu2024respiratory,
  title={Highly comparative time series analysis of oxygen saturation and heart rate to predict respiratory outcomes in extremely preterm infants},
  author={Qiu, Jiaxing and Di Fiore, Juliann M. and Krishnamurthi, Narayanan and Indic, Premananda and Carroll, John L. and Claure, Nelson and Kemp, James S. and Dennery, Phyllis A. and Ambalavanan, Namasivayam and Weese-Mayer, Debra E. and Hibbs, Anna Maria and Martin, Richard J. and Bancalari, Eduardo and Hamvas, Aaron and Moorman, J. Randall and Lake, Douglas E. and {Pre-Vent Investigators} and Krahn, Katy N. and Zimmet, Amanda M. and Hopkins, Bradley S. and Lonergan, Erin K. and Rand, Casey M. and Zadell, Arlene and Nakhmani, Arie and Carlo, Waldemar A. and Laney, Deborah and Travers, Colm P. and Vanbuskirk, Silvia and D'Ugard, Carmen and Aguilar, Ana Cecilia and Schott, Alini and Hoffmann, Julie and Linneman, Laura},
  journal={Physiological Measurement},
  volume={45},
  number={5},
  pages={055025},
  year={2024},
  doi={10.1088/1361-6579/ad4e91}
}

@inproceedings{lundberg2017shap,
  title={A unified approach to interpreting model predictions},
  author={Lundberg, Scott M. and Lee, Su-In},
  booktitle={Advances in Neural Information Processing Systems},
  volume={30},
  pages={4765--4774},
  publisher={Curran Associates, Inc.},
  year={2017}
}

@inproceedings{prokhorenkova2018catboost,
  title={CatBoost: unbiased boosting with categorical features},
  author={Prokhorenkova, Liudmila and Gusev, Gleb and Vorobev, Aleksandr and Dorogush, Anna Veronika and Gulin, Andrey},
  booktitle={Advances in Neural Information Processing Systems},
  volume={31},
  year={2018}
}

@inproceedings{chen2016xgboost,
  title={XGBoost: A scalable tree boosting system},
  author={Chen, Tianqi and Guestrin, Carlos},
  booktitle={Proceedings of the 22nd ACM SIGKDD International Conference on Knowledge Discovery and Data Mining},
  pages={785--794},
  year={2016},
  doi={10.1145/2939672.2939785}
}

@inproceedings{ke2017lightgbm,
  title={LightGBM: A highly efficient gradient boosting decision tree},
  author={Ke, Guolin and Meng, Qi and Finley, Thomas and Wang, Taifeng and Chen, Wei and Ma, Weidong and Ye, Qiwei and Liu, Tie-Yan},
  booktitle={Advances in Neural Information Processing Systems},
  volume={30},
  year={2017}
}

@article{feng2022trigger,
  title={Neonatal adverse events' trigger tool setup with random forest},
  author={Feng, Kun and Zhang, Li and He, Huayun and You, Xueqin and Zhang, Qiannan and Wei, Hong and Hua, Ziyu},
  journal={Journal of Patient Safety},
  volume={18},
  number={2},
  pages={e585--e590},
  year={2022},
  doi={10.1097/PTS.0000000000000871}
}

@book{james2021islr,
  title={An Introduction to Statistical Learning: With Applications in R},
  author={James, Gareth and Witten, Daniela and Hastie, Trevor and Tibshirani, Robert},
  edition={2},
  publisher={Springer},
  address={New York},
  year={2021},
  doi={10.1007/978-1-0716-1418-1}
}

@book{steyerberg2019prediction,
  title={Clinical Prediction Models: A Practical Approach to Development, Validation, and Updating},
  author={Steyerberg, Ewout W.},
  edition={2},
  publisher={Springer International Publishing},
  address={Cham},
  year={2019},
  doi={10.1007/978-3-030-16399-0}
}

@article{reyna2019physionet,
  title={Early prediction of sepsis from clinical data: the PhysioNet/Computing in Cardiology Challenge 2019},
  author={Reyna, Matthew A. and Josef, Christopher S. and Jeter, Russell and Shashikumar, Supreeth P. and Westover, M. Brandon and Nemati, Shamim and Clifford, Gari D. and Sharma, Ashish},
  journal={Critical Care Medicine},
  volume={48},
  number={2},
  pages={210--217},
  year={2020},
  doi={10.1097/CCM.0000000000004145}
}

@article{saito2015precision,
  title={The precision-recall plot is more informative than the ROC plot when evaluating binary classifiers on imbalanced datasets},
  author={Saito, Takaya and Rehmsmeier, Marc},
  journal={PLOS ONE},
  volume={10},
  number={3},
  pages={e0118432},
  year={2015},
  doi={10.1371/journal.pone.0118432}
}

@inproceedings{vaswani2017attention,
  title={Attention is All You Need},
  author={Vaswani, Ashish and Shazeer, Noam and Parmar, Niki and Uszkoreit, Jakob and Jones, Llion and Gomez, Aidan N. and Kaiser, Lukasz and Polosukhin, Illia},
  booktitle={Advances in Neural Information Processing Systems},
  volume={30},
  year={2017}
}

@inproceedings{lin2017focal,
  title={Focal Loss for Dense Object Detection},
  author={Lin, Tsung-Yi and Goyal, Priya and Girshick, Ross and He, Kaiming and Dollar, Piotr},
  booktitle={Proceedings of the IEEE International Conference on Computer Vision},
  pages={2980--2988},
  year={2017},
  doi={10.1109/ICCV.2017.324}
}

@inproceedings{cho2014gru,
  title={Learning Phrase Representations using RNN Encoder--Decoder for Statistical Machine Translation},
  author={Cho, Kyunghyun and Van Merrienboer, Bart and Gulcehre, Caglar and Bahdanau, Dzmitry and Bougares, Fethi and Schwenk, Holger and Bengio, Yoshua},
  booktitle={Proceedings of the 2014 Conference on Empirical Methods in Natural Language Processing},
  pages={1724--1734},
  year={2014},
  doi={10.3115/v1/D14-1179}
}

@inproceedings{wang2017time,
  title={Time series classification from scratch with deep neural networks: A strong baseline},
  author={Wang, Zhiguang and Yan, Weizhong and Oates, Tim},
  booktitle={2017 International joint conference on neural networks (IJCNN)},
  pages={1578--1585},
  year={2017},
  organization={IEEE}
}

@article{che2018grud,
  title={Recurrent Neural Networks for Multivariate Time Series with Missing Values},
  author={Che, Zhengping and Purushotham, Sanjay and Cho, Kyunghyun and Sontag, David and Liu, Yan},
  journal={Scientific Reports},
  volume={8},
  pages={6085},
  year={2018},
  doi={10.1038/s41598-018-24271-9}
}

@inproceedings{xu2024tnformermp,
  title={Temporal Neighboring Multi-Modal Transformer with Missingness-Aware Prompt for Hepatocellular Carcinoma Prediction},
  author={Xu, Jingwen and Zhu, Ye and Lyu, Fei and Wong, Grace Lai-Hung and Yuen, Pong C.},
  booktitle={Proceedings of Medical Image Computing and Computer Assisted Intervention -- MICCAI 2024},
  volume={15001},
  pages={79--88},
  publisher={Springer Nature Switzerland},
  year={2024}
}

@article{liu2025timmsfl,
  title={An End-to-End Model for Time Series Classification in the Presence of Missing Values},
  author={Liu, Mengna and Yao, Pengshuai and Cheng, Xu and Chen, Shengyong},
  journal={Expert Systems with Applications},
  volume={284},
  pages={127954},
  year={2025},
  doi={10.1016/j.eswa.2025.127954}
}

\end{document}